\documentclass{article}
\PassOptionsToPackage{numbers}{natbib}
\usepackage[preprint]{neurips_2026}
\usepackage[utf8]{inputenc}
\usepackage[T1]{fontenc}
\usepackage{amsmath,amsfonts,mathtools}
\usepackage{graphicx}
\usepackage{booktabs,multirow,tabularx}
\usepackage{microtype}
\usepackage{parskip}
\usepackage[font=small]{caption}
\usepackage{pifont}
\usepackage{wrapfig}
\usepackage{url}
\usepackage[hidelinks]{hyperref}

\title{Neither Adversarial Training Nor Purification: Emergent Adversarial Robustness from Oscillatory Predictive Learning}

\author{%
  Mohammed-Yassine Habibi\textsuperscript{1,2}\thanks{Equal contribution}~ \thanks{\raggedright
  Correspondence to:
  Mohammed-Yassine Habibi
  \texttt{<mohammed-yassine.habibi@inria.fr>}
  and Klea Ziu
  \texttt{<klea.ziu@mbzuai.ac.ae>}},
  Klea Ziu\textsuperscript{3}\footnotemark[1]~,
  Martin Tak\'a\v{c}\textsuperscript{3},
  Makoto Yamada\textsuperscript{4} \\
  \textbf{\textsuperscript{1}} Ecole polytechnique, IPP, France\\
  \textbf{\textsuperscript{2}} Ecole Nationale des Ponts et Chauss\'ees, IPP, France\\
  \textbf{\textsuperscript{3}} Mohamed bin Zayed University of Artificial Intelligence, Abu Dhabi, UAE\\
  \textbf{\textsuperscript{4}} Okinawa Institute of Science and Technology, Okinawa, Japan
}
\date{}

\begin{document}

\maketitle

\begin{abstract}
Adversarial robustness in computer vision is still largely achieved through adversarial training or test-time adversarial purification, both of which introduce significant computational overhead by generating adversarial examples during training or performing iterative denoising at test time. We study whether empirical robustness can instead emerge from architectural and representation-learning inductive biases. We introduce Oscillatory Predictive Learning (OPL), a two-stage framework that combines Artificial Kuramoto Oscillatory Neurons (AKOrN) with predictive self-supervised pretraining using X-PhiNet. Because our default checkpoint uses randomized initial oscillator states, we compare it with other randomized adversarial defense methods that provide precise, reproducible, and strong attack protocols. Experiments on CIFAR-10 and CIFAR-100, with additional corruption evaluation on CIFAR-10-C, demonstrate that our method achieves competitive results under the AutoAttack-rand evaluation protocol. On CIFAR-10 and CIFAR-100, OPL attains 76.63\(\pm\)0.76\% and 50.44\% robust accuracy, respectively, under \(\ell_\infty\), \(\epsilon=8/255\), AutoAttack-rand with EoT \(K=20\). In addition, we conduct several diagnostics for attack failure and gradient masking via black-box attacks, transfer attacks, and attack-strength sweeps. Ablation studies show that oscillatory dynamics provide a key robustness-inducing inductive bias, which is significantly amplified by predictive pretraining. We further find that robustness is concentrated in a narrow dynamical regime, with changes in oscillator dimension sharply reducing robust accuracy despite improving clean accuracy. Our results suggest that architectural and self-supervised predictive biases can offer an efficient and complementary pathway to empirical adversarial robustness, challenging the view that high robustness necessarily requires adversarial-example generation during training or iterative purification at test time.
\end{abstract}

\section{Introduction}

Deep vision models remain vulnerable to small adversarial perturbations \cite{Szegedy2013_intriguing, bartoldson2024adversarial}, and the strongest widely adopted robustness results are still dominated by adversarial training \cite{amini_meansparse_2024,bartoldson2024adversarial, kang_stable_2021} and adversarial purification \cite{nie_diffusion_nodate, croce_evaluating_2022}. Indeed, there are no strong baselines for vision models that rely on neither adversarial training nor adversarial purifying \cite{croce_robustbench_2021, croce_evaluating_2022}. Although effective, those methods substantially increase training costs because they require generating adversarial examples throughout optimization or performing iterative denoising at test time, often at scale. This computational burden is not merely an engineering inconvenience: recent scaling analyses argue that, under current adversarial-training paradigms, simply increasing compute is unlikely to close the gap to human-level robustness, and that meaningful progress will require more efficient training algorithms and improved architectures rather than scale alone \cite{bartoldson2024adversarial}. This motivates the question explored: can adversarial robustness emerge from architectural and representation-learning inductive biases rather than from repeated exposure to adversarial examples?

In this paper, we test a concrete hypothesis: a predictive self-supervised objective combined with oscillatory encoder dynamics can yield strong robustness under standardized adversarial evaluation, without adversarial training. We study this hypothesis through \textbf{Oscillatory Predictive Learning}, a framework that combines an Artificial Kuramoto Oscillatory Neurons (AKOrN) encoder \cite{miyato2025artificial} with predictive self-supervised learning via X-PhiNet \cite{ishikawa2025phinets}. Oscillatory coupling imposes structured consensus dynamics that can reduce sensitivity to local perturbations, while predictive pretraining encourages stable and globally consistent representations in this dynamical setting. The biological analogy motivates this design, but the contribution we evaluate is algorithmic: the robustness arises from this interaction, and the combination of AKOrN and X-PhiNet substantially outperforms either component alone under standardized attacks.

To assess whether the robustness gain is specific to X-PhiNet or reflects a broader benefit of predictive non-contrastive pretraining, we compare several SSL objectives on the same AKOrN backbone in Appendix \ref{app:f_exp_res}. In these preliminary experiments, X-PhiNet gives the strongest robust accuracy among the tested methods, while SimSiam~\cite{chen_exploring_2021} achieves a similar but slightly lower result, and BYOL~\cite{grill_bootstrap_2020} is less stable. We therefore use X-PhiNet as the main pretraining objective in the rest of the paper. We do not claim that the effect automatically transfers to all predictive SSL frameworks; evaluating JEPA-style~\cite{destrade_value-guided_2025, assran_self-supervised_2023} objectives and larger predictive architectures remains future work.

Empirically, this is what we observe. On CIFAR-10, our method, \textbf{Oscillatory Predictive Learning}, achieves $76.63\pm 0.76\%$ robust accuracy under \(\ell_\infty\), \(\epsilon=8/255\), AutoAttack-rand with EoT \(K=20\) \cite{croce_reliable_adv_rob_2020, croce_evaluating_2022}, outperforming prior stochastic defenses evaluated under the same AutoAttack-rand protocol that has been recommended by \cite{croce_evaluating_2022, lee_robust_2023} (\textit{DiffPure}: 71.29\%, \cite{nie_diffusion_nodate}) . In the same evaluation, AKOrN alone reaches 64.98\% AutoAttack-rand robustness, while X-PhiNet alone and a standard ResNet baseline remain almost entirely non-robust. This pattern supports a complementary view of the two components: oscillatory encoder dynamics provide a robustness-supporting inductive bias, and predictive pretraining substantially amplifies it.

A central outcome of this discovery is computational. Because our method does not rely on adversarial training, it avoids the dominant source of cost in current robust-learning pipelines. As a result, it offers a substantially lower-compute route to robustness and a more promising starting point for scaling than methods whose training procedure already embeds adversarial-example generation at every iteration. In our current experiments, the robustness is achieved at the cost of lower clean accuracy compared to high-capacity adversarially trained models. Beyond aggregate performance, we find a sharp sensitivity to oscillator dimension. In our sweep, robustness is concentrated in the \(N=2\) family, while increasing the oscillator dimension to \(N=4\) substantially reduces robust accuracy despite improving clean accuracy. Increasing encoder flexibility or optimizing for clean accuracy does not monotonically improve adversarial robustness; in several settings, it actively degrades it. This suggests that robustness in oscillatory models is not simply a byproduct of larger capacity or better standard representations, but depends on maintaining specific dynamical constraints.

Together, these findings point to a complementary route to adversarial robustness: instead of explicitly fitting adversarial perturbations during training, one can induce robustness through the interaction between structured encoder dynamics and predictive learning. We therefore propose \textbf{Oscillatory Predictive Learning} as a framework for studying robustness from inductive bias, and we evaluate it through standardized attacks, controlled ablations, and dynamical hyperparameter sweeps that expose a reproducible robust regime.

\paragraph{Contributions}
In summary, our work provides a highly efficient alternative to adversarial optimization through the following principal contributions:
\begin{itemize}
    \item We propose \textbf{Oscillatory Predictive Learning}, a framework combining AKOrN encoders with X-PhiNet predictive pretraining, achieving 76.63\% and 50.44\% AutoAttack-rand robustness on CIFAR-10 and CIFAR-100, respectively, entirely without adversarial training or adversarial purification.
    \item We report OPL in a substantially lower measured training-FLOP regime than representative adversarially trained CIFAR models. Because the compared systems differ in architecture, data, training procedure, and evaluation protocol, we treat these results as contextual rather than as a protocol-matched efficiency comparison.
    \item We provide controlled evidence that oscillatory dynamics and predictive pretraining play complementary roles: X-PhiNet provides a substantial +11.65 percentage point robustness amplification when applied to AKOrN (from 64.98\% to 76.63\%), whereas applying X-PhiNet to a standard ResNet yields near-zero robustness.
    \item We identify a  \emph{sharp sensitivity to oscillator dimensionality} driven by dynamical hyperparameters, demonstrating that robustness collapses near-discontinuously when increasing oscillator dimensionality (e.g., from $N=2$ to $N=4$), revealing that clean and robust accuracy are not monotonically aligned.
\end{itemize}

\section{Background and Motivation}

\paragraph{Standardized evaluation of adversarial robustness.}
Empirical robustness claims depend strongly on the attack used for evaluation. Deterministic CIFAR classifiers are commonly evaluated with standard AutoAttack and reported on RobustBench. Since OPL uses randomized initial oscillator states, we instead evaluate it as a stochastic classifier using AutoAttack-rand with EoT. We therefore report deterministic RobustBench-style results only as contextual references, not as protocol-matched comparisons.\cite{croce_robustbench_2021,croce_reliable_adv_rob_2020}. We report additional $l_{\infty}$ and $\ell_2$ robust accuracy and corruption/noise evaluations as complementary stress.

\paragraph{Adversarial training at scale.}
One of the two strongest CIFAR robustness methods is adversarial training, often combined with large synthetic datasets and high-capacity backbones \cite{bartoldson2024adversarial,amini_meansparse_2024}. This line of work is highly effective, but its cost profile is increasingly extreme. As emphasized in \ \cite{bartoldson2024adversarial}, adversarially trained CIFAR-10 models now operate in a regime ranging from tens of millions of synthetic examples to hundreds of millions, culminating in a 300M-unique-sample / 500M-sample training pipeline and over $10^{21}$ training FLOPs \cite{bartoldson2024adversarial, amini_meansparse_2024}.

\paragraph{Dynamical models for robustness.}
Prior work has linked robustness to dynamical stability. Stable neural ODEs impose Lyapunov-stable equilibrium behavior so that small perturbations are driven toward similar trajectories \cite{kang_stable_2021}. AKOrN extends this dynamical viewpoint by replacing static threshold units with coupled Kuramoto-style oscillator updates \cite{miyato2025artificial}. In this paper, we treat oscillatory dynamics as a robustness-supporting inductive bias.

\paragraph{Stochastic and purification-based defenses.}
OPL belongs to a broader class of stochastic defenses that leverage randomness at inference time to improve adversarial robustness. This includes both randomized smoothing methods \cite{cohen_certied_nodate}, which inject noise in input space to obtain certified guarantees, and adversarial purification approaches, such as diffusion-based defenses, which iteratively denoise perturbed inputs through stochastic generative processes before classification. These methods typically require adaptive evaluation via Expectation over Transformation (EoT) \cite{athalye_obfuscated_2018} to account for inference-time randomness \cite{croce_evaluating_2022}. In contrast, OPL does not rely on explicit input perturbation or generative purification; instead, stochasticity arises intrinsically from the random initialization of the internal oscillators. This yields strong empirical robustness without adversarial training, noise augmentation, or iterative test-time preprocessing.

\paragraph{Self-supervised approaches to robustness.}
Prior work have demonstrated that self-supervised learning can improve model robustness and uncertainty estimation \cite{hendrycks_using_nodate}, particularly as a pre-training step for adversarial fine-tuning \cite{chen_adversarial_2020}. Predictive learning serves as a general non-contrastive SSL framework, prominently featured in methods like SimSiam \cite{chen_exploring_2021} and BYOL \cite{grill_bootstrap_2020}, which learn stable representations by predicting augmented views. While these methods rely on general architectural constraints to prevent collapse, X-PhiNet \cite{ishikawa2025phinets} makes an explicit link to biological temporal prediction. We build upon this foundation, showing that predictive SSL without adversarial fine-tuning yields state-of-the-art robustness when paired with the right dynamical architecture.

\paragraph{Recurrent and iterative-inference defenses.}
Iterative inference mechanisms, such as deep equilibrium networks and recurrent reasoning steps, naturally resist adversarial perturbations by driving activations toward stable attractors \cite{chu_lyapunov-stable_2024}. OPL extends this philosophy by operating as a continuous-time dynamical system via Kuramoto coupling \cite{miyato2025artificial}.

Our work sits at the intersection of these directions. Unlike prior top-performing defenses, we do not use adversarial examples during training or adversarial purification during test-time inference. Unlike prior dynamical or self-supervised methods considered in isolation, we study robustness as an interaction between architecture and objective: the encoder provides the oscillatory inductive bias while the specific SSL framework provides the predictive learning signal, and their combination yields strong standardized robustness in a narrow dynamical regime.

\section{Preliminaries: Oscillatory Encoders and Predictive Pretraining}

\paragraph{AKOrN as a coupled oscillatory encoder.}
Let $x \in \mathbb{R}^{3 \times H_0 \times W_0}$ denote an input image. A convolutional stem maps $x$ to a feature map
$X \in \mathbb{R}^{C \times H \times W}$ with $C = KN$ channels. We reshape this tensor as
$X \in \mathbb{R}^{K \times N \times H \times W}$ so that each spatial location $(h,w)$ contains $K$ oscillator states
$\mathbf{x}_{k,h,w} \in \mathbb{S}^{N-1}$, where
\[
\mathbb{S}^{N-1} \coloneqq \{ \mathbf{u} \in \mathbb{R}^N : \|\mathbf{u}\|_2 = 1 \}.
\]
Here, $N$ is the oscillator dimension (the number of rotating dimensions), and $K=C/N$ is the number of oscillators per spatial location.

A Kuramoto layer updates each oscillator through a coupled dynamical system:
\begin{equation}
\dot{\mathbf{x}}_{k,h,w}
=
\mathbf{\Omega}_{k,h,w}\mathbf{x}_{k,h,w}
+
\mathrm{Proj}_{\mathbf{x}_{k,h,w}}
\left(
\mathbf{y}_{k,h,w}
+
\sum_{k',\Delta h,\Delta w}
\mathbf{J}_{k,k',\Delta h,\Delta w}\,
\mathbf{x}_{k',h+\Delta h,w+\Delta w}
\right),
\label{eq:theoretical_akorn}
\end{equation}
where $\mathbf{\Omega}_{k,h,w}\in\mathbb{R}^{N\times N}$ is a learned antisymmetric matrix controlling intrinsic rotation,
$\mathbf{J}_{k,k',\Delta h,\Delta w}$ are learned coupling weights,
and $\mathbf{y}_{k,h,w}$ is an input-dependent drive term. Because we use an implementation of the Kuramoto dynamical system with convolutions in our experiments, $\mathbf{J}_{k,k',\Delta h,\Delta w}$ will be the learned connectivity weights of a convolution kernel and $k, k'$ stand for output and input channels indexes.
The projection operator
\[
\mathrm{Proj}_{\mathbf{x}}(\mathbf{y}) = \mathbf{y} - \langle \mathbf{y}, \mathbf{x} \rangle \mathbf{x}
\]
ensures the velocity vector of an oscillator $\mathbf{x}$ effectively lives in the tangent plane of $\mathbf{S}^{N-1}$ at $\mathbf{x}$.

The coupling term is the key inductive bias of AKOrN. Each oscillator is updated using both its own state and the states of neighboring oscillators, which encourages locally consistent configurations in feature space. Intuitively, if a perturbation affects only a subset of local states, the coupled dynamics need not amplify those deviations independently; instead, neighboring oscillators can pull the representation back toward a consistent configuration.

In practice, Eq.~\eqref{eq:theoretical_akorn} is implemented by unrolling the dynamics for $T$ discrete steps within each of $L$ Oscillatory Neurons Units. Full derivations and implementation details are deferred to Appendix~\ref{app:Maths}.

\paragraph{Predictive self-supervised pretraining.}
 X-PhiNet provides the pretraining objective used to shape the encoder before supervised fine-tuning. At a high level, X-PhiNet is a non-contrastive learning protocol that learns by predicting one transformed view from another, rather than by contrasting each sample against large sets of negatives \cite{ishikawa2025phinets}. Its neuroscience motivation comes from the temporal prediction hypothesis: a CA3-like predictor introduces a short delay between signals, and a CA1-like predictor learns to compensate that delay. In self-supervised learning terms, this can be viewed as encouraging representations that remain predictive across nearby observations of the same underlying scene while preserving semantically relevant structure.

This predictive objective is also a natural candidate for robustness-oriented representation shaping. Predictive self-supervision encourages invariances across transformed views of the same input and can promote stable latent structure, which may reduce sensitivity to small perturbations, although it does not guarantee adversarial robustness by itself. In our setting, predictive pretraining therefore acts as a complementary inductive bias to oscillatory dynamics. AKOrN constrains \emph{how} features evolve through the encoder, while X-PhiNet constrains \emph{which} representations are learned. Our central question is whether robustness emerges specifically from this interaction, rather than from either ingredient alone.

As shown in Figure \ref{fig:phinet} below, X-PhiNet, a single input image $x$ is augmented in two different ways, resulting in three inputs $x$, $x^{(1)}$, $x^{(2)}$ that are processed by three parallel streams. The first two streams each contain an encoder $f$ while the last one contains a different encoder $f_{long}$.

\clearpage

\begin{wrapfigure}{r}{0.42\textwidth}
    \vspace{-1.3\baselineskip} %
    \centering
    \includegraphics[width=\linewidth]{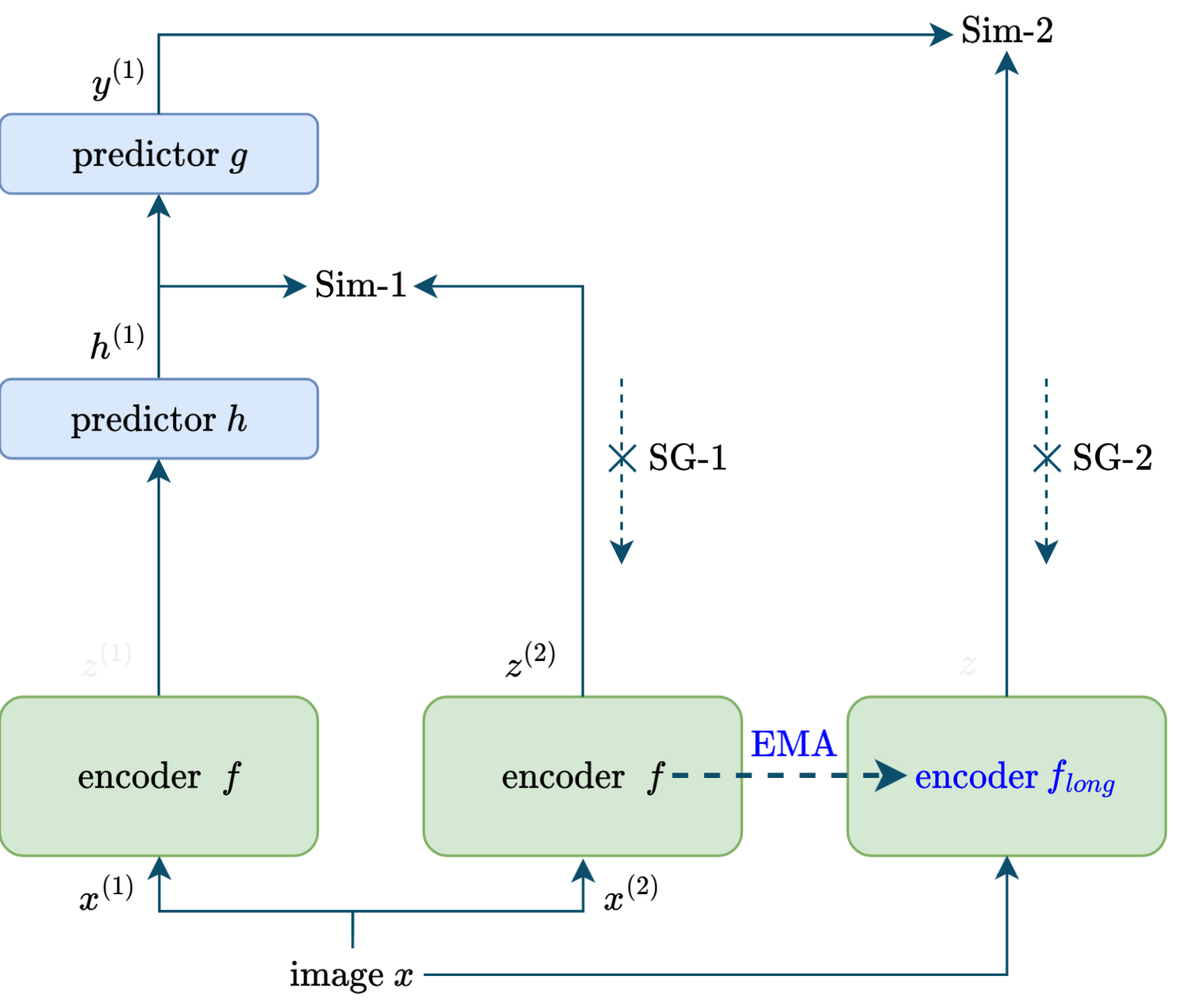}
    \caption{X-PhiNet Temporal Prediction Framework.}
    \label{fig:phinet}
\end{wrapfigure}

 The outputs of these streams are compared at different depths by computing their similarities. The model is trained by jointly minimizing two losses: the first one, Sim-1, is a symmetric negative cosine loss function for the hippocampus model, and the second one, Sim-2, is a Mean Squared Error loss, which is a slow-learning loss for the neocortex model. Those two losses are combined to update the encoder $f$ by backpropagation. The third stream is designed to model the neocortex by $f_{long}$, which corresponds to a stable encoder that further improves slow learning with an ability to maintain long-term signals. In our experiments, we extract the $f_{long}$ once the pretraining is complete.

 We give the exact loss used in our implementation in Appendix \ref{app:Maths} and summarize the overall training pipeline in Section~\ref{sec:experiments}.

\section{Oscillatory Predictive Learning}
\paragraph{Overview.}
The Oscillatory Predictive Learning framework explored here combines an AKOrN encoder with X-PhiNet pretraining in a two-stage pipeline.

In the first stage, we pretrain the AKOrN encoder using the X-PhiNet objective to learn view-consistent oscillatory representations. In the second stage, we replace the pretraining heads with a classification head and fine-tune the model for supervised image classification. This design lets us test the central hypothesis of the paper directly: whether predictive self-supervision amplifies the robustness bias induced by oscillatory encoder dynamics.

\paragraph{Architecture.} The encoder consists of a convolutional stem followed by $L$ Oscillatory Neurons Units blocks, each unrolled for $T$ discrete dynamics steps. Let $C$ denote the total channel width and $N$ the oscillator dimension; then each spatial location contains $K=C/N$ oscillators. After the final Oscillatory Neurons Unit, global pooling produces a representation that is fed either to the X-PhiNet pretraining heads or to a supervised classification head. The initial oscillatory state $X^{in}$ for the first layer of the model can be chosen randomly or deterministically. We evaluate both stochastic and deterministic variants of OPL.
For stochastic models, we use AutoAttack-rand with Expectation over Transformation (EoT) to correctly estimate gradients under randomness. For deterministic models, the forward pass is fixed, and EoT has no effect. \cite{athalye_obfuscated_2018, croce_reliable_adv_rob_2020}.

\begin{figure}[htbp]

    \centering

    \includegraphics[width=0.6\textwidth]{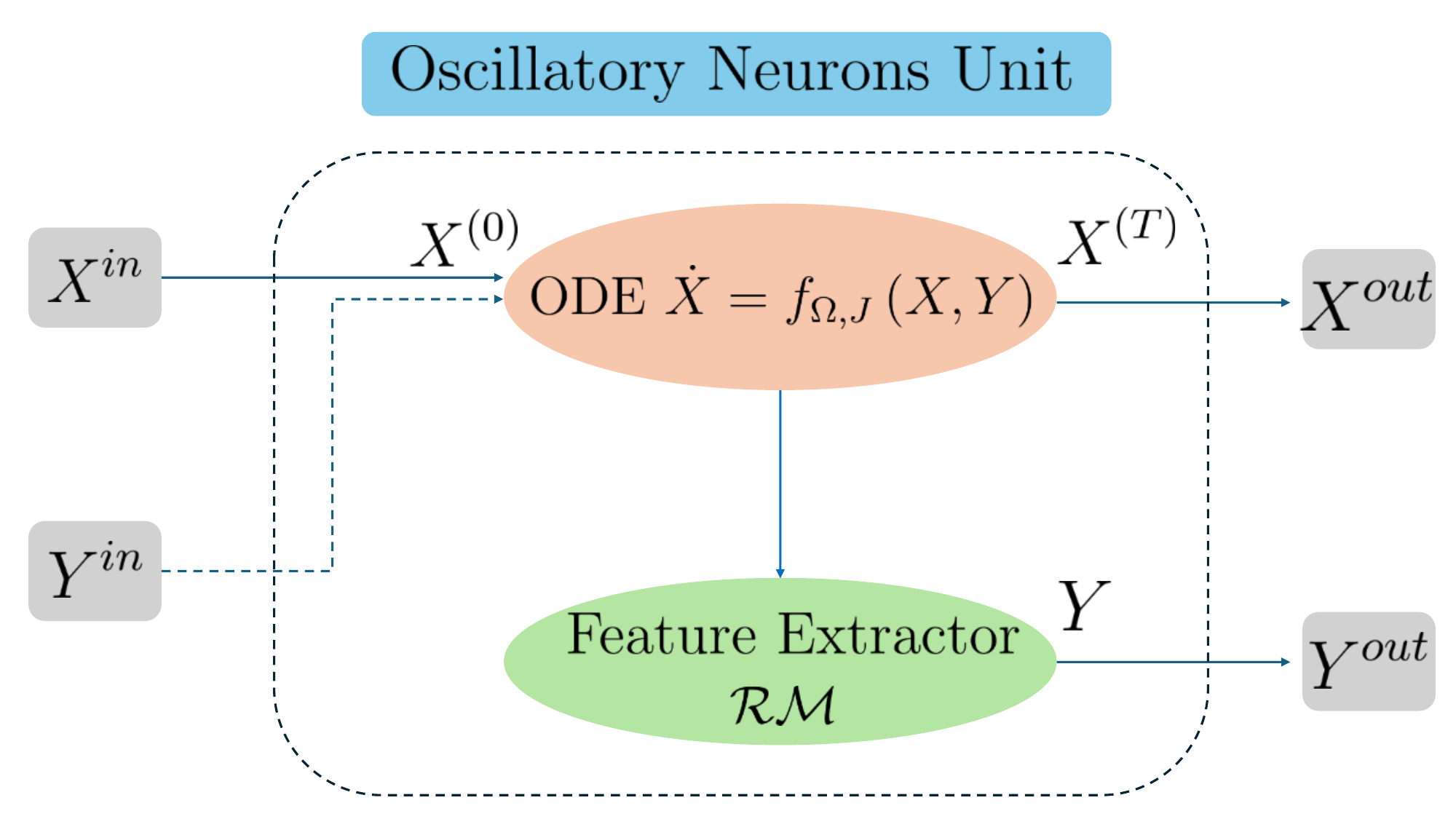}

    \caption{Stream of a single Oscillatory Neurons Unit used in the encoder architecture. The inputs are an initial oscillatory state $X^{in}\in \mathbb{R}^{K \times N \times H \times W}$, and a conditional stimuli (in practice the input image) $Y^{in}\in \mathbb{R}^{K \times N \times H \times W}$. The outputs are the final oscillatory state $X^{out}$ and the extracted features $Y^{out}$.}

    \label{fig:oscillatory_unit}

\end{figure}

\paragraph{Training pipeline.}
During pretraining, we optimize the X-PhiNet objective using the AKOrN encoder and the corresponding pretraining heads required by the predictive learning setup. After pretraining, these heads are discarded and replaced with a classification head. We then fine-tune the resulting model on labeled training data using standard cross-entropy loss. Unless otherwise stated, the encoder architecture is unchanged between the two stages.

\paragraph{Notation.}
We denote a model configuration by $(C, N, T, L)$, where $C$ is the channel width, $N$ is the oscillator dimension, $T$ is the number of unrolled dynamics steps per block, and $L$ is the number of Oscillatory Neurons Units. The default configuration used in our experiments is reported in Section~\ref{sec:experiments}.

\section{Experiments}
\label{sec:experiments}

We evaluate OPL on \textsc{CIFAR-10}, CIFAR-10-C and \textsc{CIFAR-100} under $\ell_\infty$ and $\ell_2$ threat models.
Because the checkpoint studied in this paper uses randomized initial oscillator states, our primary robustness evaluation uses \textbf{AutoAttack-rand} with Expectation over Transformation (EoT). We primarily compare our result with other SOTA methods reporting \textbf{AutoAttack-rand} with Expectation over Transformation (EoT) results.
We additionally report PGD, Square Attack, transfer attacks, and Gaussian-noise robustness as complementary diagnostics.

\subsection{Experimental Setup}
\label{sec:setup}

\paragraph{Architecture and training.}
Our default OPL encoder uses $C=128$ channels, $N=2$ oscillator dimensions, $L=3$ Oscillatory Neurons Units, and $T=3$ integration steps per unit.
We pre-train the encoder with the X-PhiNet temporal prediction framework for 400 epochs and fine-tune it for 400 epochs on the same dataset (denoted \textbf{400/400}).
Unless otherwise stated, all OPL results in this section use this configuration.
Full implementation details, including optimizer, learning-rate schedule, and EMA coefficients, are provided in Appendix~\ref{app:exp_sett} and \ref{app:f_exp_res}.

\paragraph{Attack suite.}
We report robustness under five evaluation protocols.

\begin{itemize}
    \item \textbf{AutoAttack-rand}~\citep{croce_reliable_adv_rob_2020}: our primary evaluation for the randomized OPL checkpoint. We use $\ell_\infty$ attacks with $\varepsilon=8/255$ and EoT with $K=20$ samples. When available, we report mean $\pm$ std over five independently trained models.

    \item \textbf{PGD} ($\ell_\infty$ and $\ell_2$): we use $\varepsilon=8/255$ and $\varepsilon=0.5$, respectively, with 5 random restarts and a step-count sweep $\{10,20,50,100\}$. For randomized checkpoints, PGD uses EoT with $K=5$. We use these attacks as stronger first-order diagnostics rather than as a replacement for AutoAttack.

    \item \textbf{Square Attack}~\citep{andriushchenko2020square}: a black-box, score-based attack used as a complementary diagnostic that does not rely on backpropagated gradients.

    \item \textbf{Transfer attacks}: PGD adversarial examples are generated on a surrogate ResNet-50 and then evaluated on the target model.

    \item \textbf{Gaussian-noise robustness}: we measure accuracy under increasing additive Gaussian noise to assess stability to non-adversarial corruptions.
\end{itemize}

All $\ell_\infty$ evaluations use $\varepsilon=8/255$ on CIFAR-10/100.
Where relevant, we include published RobustBench numbers only as context, because those references are reported under the standard AutoAttack protocol and are not directly protocol-matched to randomized OPL.

\subsection{Comprehensive Adversarial Evaluation}
\label{sec:eval}

\paragraph{Main result.}
Table~\ref{tab:robustbench} reports the adaptive robustness of randomized OPL under AutoAttack-rand with EoT ($K=20$).
Without adversarial training, OPL reaches \textbf{76.63\% $\pm$ 0.76\%} robust accuracy on CIFAR-10 and \textbf{50.44\%} on CIFAR-100.
Table~2 provides contextual comparison with prior stochastic or purification-based defenses evaluated using EoT-style adaptive protocols. Because these methods differ in architecture, test-time computation, and exact attack implementation, we use this comparison to position OPL. Best reported robust accuracy on stochastic adversarial defense models on CIFAR-10 under the same AutoAttack-rand protocol is 71.29\%, meaning we claim $\mathbf{+5.34\%}$.
All OPL results in this subsection use the same 400/400 training setup described in Section~\ref{sec:setup}.
\begin{table}[h]
\centering
\small
\caption{
\textbf{Adaptive white-box evaluation of randomized OPL.}
OPL is evaluated with AutoAttack-rand using EoT ($K=20$).
}
\label{tab:robustbench}
\begin{tabular}{lcc}
\toprule
\textbf{Dataset} & \textbf{Evaluation protocol} & \textbf{Robust accuracy (\%)} \\
\midrule
\textsc{CIFAR-10}  & AA-rand, $\ell_\infty$, $\varepsilon=8/255$, EoT $K=20$ & $\mathbf{76.63 \pm 0.76}$ \\
\textsc{CIFAR-100} & AA-rand, $\ell_\infty$, $\varepsilon=8/255$, EoT $K=20$ & $\mathbf{50.44}$ \\
\bottomrule
\end{tabular}
\end{table}
\begin{table}[h]
\centering
\small
\caption{
\textbf{Robustness to AutoAttack-rand adversarial attack} Contextual comparison with stochastic or adaptive defenses on CIFAR-10 under \(\ell_\infty\), \(\epsilon=8/255\). OPL is evaluated with AutoAttack-rand and EoT \(K=20\) (Standard \cite{croce_robustbench_2021} evaluation scheme).}
\label{tab:rob_autoattack_rand}
\begin{tabular}{lc}
\toprule
\textbf{Model} & \textbf{AutoAttack-rand robust acc. ($\%$)}\\
\midrule
Nie et al. (2022) \cite{nie_diffusion_nodate}& 71.29\\
Lee et al. (2023) \cite{lee_robust_2023}& 70.47\\
AKOrN \cite{miyato2025artificial}& 64.98 \\
\textbf{OPL (ours)} & \textbf{76.63 $\pm$ 0.76} \\
\bottomrule
\end{tabular}

\end{table}

\subsection{Adaptive Evaluation Diagnostics}
\label{sec:gm}
Stochastic defenses are vulnerable to overestimated robustness if attacks do not properly account for randomness or iterative computation. We therefore report several diagnostics: AutoAttack-rand with EoT, PGD attack-strength sweeps, Square Attack, and transfer attacks. These diagnostics test several common failure modes.

\paragraph{PGD robustness.}
Table~\ref{tab:pgd} reports PGD accuracy as a function of attack iterations.
Robust accuracy decreases monotonically as the attack budget increases, indicating that stronger first-order attacks continue to find additional adversarial examples rather than plateauing prematurely.
Under $\ell_\infty$ PGD-100, OPL reaches 76.10\% robust accuracy; under $\ell_2$ PGD-100, it reaches 82.11\%.

\begin{table}[h]
\centering
\small
\caption{
\textbf{PGD robustness on randomized OPL for CIFAR-10.}
Clean accuracy: 86.81\%.
PGD uses 5 restarts and EoT ($K=5$).
The monotonic decrease with attack strength is consistent with informative gradients.
}
\label{tab:pgd}
\begin{tabular}{lcccc}
\toprule
\textbf{Attack} & \textbf{10} & \textbf{20} & \textbf{50} & \textbf{100} \\
\midrule
$\ell_2$ ($\varepsilon = 0.5$)        & 82.32 & 82.28 & 82.26 & 82.11 \\
$\ell_\infty$ ($\varepsilon = 8/255$) & 79.80 & 77.90 & 76.51 & 76.10 \\
\bottomrule
\end{tabular}
\vspace{0.35em}

\end{table}

\paragraph{Square Attack.}
Under Square Attack (black-box, score-based, $\ell_\infty$), OPL achieves \textbf{81.81\%} robust accuracy from a clean accuracy of 86.81\%.
Because Square Attack does not rely on backpropagated gradients, it provides a complementary diagnostic to the white-box evaluations.
Notably, Square Attack achieves higher accuracy (81.81\%) than AutoAttack-rand (76.63\%), which is the expected ordering when gradients are informative; the opposite ordering is a canonical signature of gradient masking.

\paragraph{Transfer attacks.}
Table~\ref{tab:transfer} reports robustness to PGD adversarial examples crafted on a ResNet-50 surrogate.
We treat transfer attacks as a compementary evaluation to adaptive white-box evaluations.
OPL is more resistant than AKOrN alone under both \(\ell_\infty\) and \(\ell_2\) transfer attacks, suggesting that predictive pretraining improves off-model stability.  However, the \(\ell_\infty\) transfer result for OPL is lower than the AutoAttack-rand result. We therefore do not interpret AutoAttack-rand as a worst-case lower bound across all possible attacks; rather, transfer attacks provide an additional diagnostic indicating that stronger adaptive attacks may further reduce robust accuracy.

\begin{table}[h]
\centering
\small
\caption{
\textbf{Transfer attack (ResNet-50 $\rightarrow$ target) on CIFAR-10.}
PGD-100 adversarial examples are generated on the surrogate model and evaluated on the target model.
The $\ell_\infty$ and $\ell_2$ columns use $\varepsilon=8/255$ and $\varepsilon=0.5$, respectively.
}
\label{tab:transfer}
\begin{tabular}{lccc}
\toprule
\textbf{Model} & \textbf{Clean (\%)} & \textbf{PGD-$\ell_\infty$ (\%)} & \textbf{PGD-$\ell_2$ (\%)} \\
\midrule
AKOrN          & 84.99 & 67.05 & 76.32 \\
OPL & 86.81 & 70.59 & 78.52 \\
\bottomrule
\end{tabular}
\vspace{0.35em}

\end{table}
\subsection{Ablation Studies}
\label{sec:ablations}

We use ablations to answer two questions: (i) does robustness primarily come from the oscillatory encoder or from predictive pretraining, and (ii) which dynamical hyperparameters place the model in a robust or collapsed regime?
Unless otherwise stated, robust accuracy in this subsection is measured with \textbf{AA-rand} ($\ell_\infty$, $\varepsilon=8/255$, EoT $K=20$), matching the primary evaluation in Section~\ref{sec:eval}.
We report PGD diagnostics for the full model separately in Table~\ref{tab:pgd} and avoid repeating them here.

\subsubsection{Architecture vs.\ Predictive Pretraining}
\label{sec:ablation_arch}

Table~\ref{tab:main} isolates four conditions:
(i) a standard ResNet-50 baseline,
(ii) X-PhiNet pretraining on ResNet-50,
(iii) AKOrN without predictive pretraining, and
(iv) the full OPL model.
The pattern is clear.
Applying X-PhiNet to a standard ResNet-50 encoder does not yield measurable adversarial robustness on its own (0.31\%), whereas AKOrN without predictive pretraining is already substantially robust (64.98\%). Within the tested model family, this indicates that oscillatory dynamics are the primary robustness-supporting component, while predictive pretraining alone is insufficient on a standard ResNet encoder.
Moreover, adding the X-PhiNet predictive pretraining pipeline on top of AKOrN increases robust accuracy to 76.63 $\pm$ 0.76\%, a gain of 11.65 percentage points over AKOrN alone.
This result suggests that predictive pretraining substantially strengthens the robustness bias induced by the oscillatory encoder AKOrN.

\begin{table}[h]
\caption{
\textbf{Disentangling architecture and predictive pretraining on CIFAR-10.}
All models are evaluated under the same randomized white-box protocol: AA-rand ($\ell_\infty$, $\varepsilon=8/255$, EoT $K=20$).
The table isolates the contributions of a standard encoder, predictive pretraining, oscillatory dynamics, and their combination.
}
\label{tab:main}
\centering
\small
\setlength{\tabcolsep}{5pt}
\begin{tabular}{lcccc}
\toprule
\textbf{Model} & \textbf{Oscillatory encoder} & \textbf{Predictive SSL} & \textbf{Clean (\%)} & \textbf{AA-rand-$\ell_\infty$ (\%)} \\
\midrule
ResNet-50                & No  & No  & 82.17 & 0.31 \\
 X-PhiNet + ResNet-50     & No  & Yes & 82.04 & 0.17 \\
AKOrN                    & Yes & No  & 84.99 & 64.98 \\
\midrule
\textbf{OPL} & \textbf{Yes} & \textbf{Yes} & \textbf{86.81} & \textbf{76.63 $\pm$ 0.76} \\
\bottomrule
\end{tabular}
\end{table}

\begin{wrapfigure}{r}{0.48\textwidth}
    \centering
     \vspace{-1pt}
\includegraphics[width=0.46\textwidth]{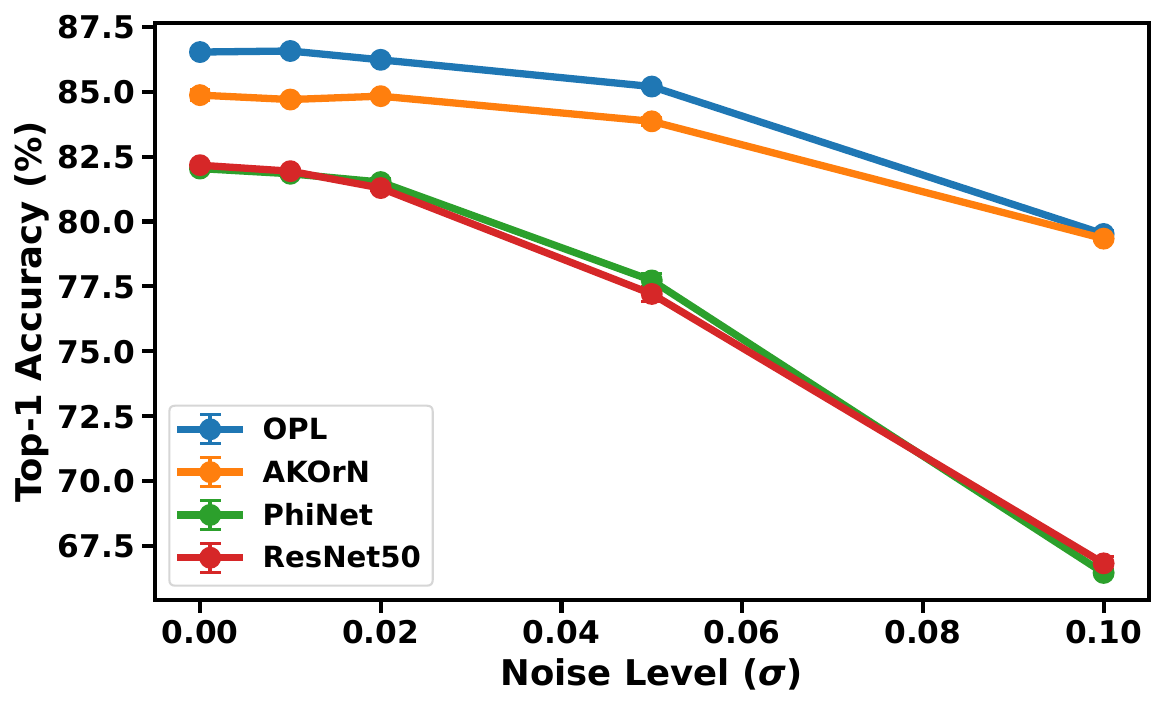}
    \caption{
    \textbf{Robustness to additive Gaussian noise on CIFAR-10.}
    We report mean top-1 test accuracy as a function of the Gaussian noise standard deviation $\sigma$.
    Each point is averaged over three independently sampled Gaussian-noise draws on the full 10{,}000-image test set; $\sigma=0$ corresponds to clean accuracy.
    OPL maintains the highest accuracy across the evaluated noise range, while AKOrN alone remains more stable than the X-PhiNet-only and ResNet-50 controls.
    }
    \label{fig:gaussian}
    \vspace{-1.5\baselineskip}
\end{wrapfigure}

\textbf{Gaussian-noise robustness.}
Figure~\ref{fig:gaussian} shows that the same ordering extends beyond adversarial perturbations.
AKOrN is consistently more stable than the ResNet-50 and X-PhiNet-only controls across the full noise range, and OPL performs best throughout.
This result does not replace adversarial evaluation, but it supports the broader picture that oscillatory dynamics improve stability to both worst-case and stochastic perturbations.

\subsubsection{Hyperparameter sensitivity and the robust regime.}
\label{sec:ablation_hparams}

Table~\ref{tab:hparam} summarizes the sweep over oscillator dimension $N$, integration time $T$, and training schedule.
Two regimes emerge.
The $N{=}2$ family contains all robust configurations, with the 400/400, $T{=}3$ model yielding the strongest overall trade-off.
Within this family, increasing $T$ from 3 to 5 sharply reduces clean accuracy while also reducing robust accuracy, so longer unrolling does not improve robustness in our setting.
By contrast, increasing oscillator dimension from $N{=}2$ to $N{=}4$ raises clean accuracy toward 92\% but collapses AA-rand robustness to near zero.
This non-monotonic pattern argues against a simple capacity explanation: robustness appears only in a narrow dynamical regime rather than improving monotonically with model flexibility.

\begin{table}[h]
\caption{
\textbf{Hyperparameter sensitivity on CIFAR-10.}
One representative result is shown for each unique configuration in the current sweep.
All robust accuracies are measured with AA-rand ($\ell_\infty$, $\varepsilon=8/255$, EoT $K=20$).
Channel width is fixed at $C=128$ throughout.
Robustness is concentrated in the $N{=}2$ family; increasing oscillator dimension to $N{=}4$ improves clean accuracy but collapses robust accuracy.
}
\label{tab:hparam}
\centering
\small
\setlength{\tabcolsep}{5pt}
\begin{tabular}{ccccc c}
\toprule
\textbf{Pre.} & \textbf{Fine.} & \boldmath$N$ & \boldmath$T$ & \boldmath$L$ & \textbf{Clean (\%) -- Robust (\%)} \\
\midrule
400 & 400 & 2 & 3 & 3 & \textbf{86.81 -- 76.63} \\
200 & 400 & 2 & 3 & 3 & 84.68 -- 69.56 \\
400 & 400 & 2 & 5 & 3 & 65.36 -- 54.05 \\
50  & 50  & 2 & 5 & 3 & 53.56 -- 46.54 \\
\midrule
400 & 400 & 4 & 3 & 3 & 92.06 -- 0.03 \\
400 & 100 & 4 & 3 & 3 & 81.65 -- 4.01 \\
400 & 400 & 4 & 5 & 3 & 92.10 -- 0.02 \\
50  & 50  & 4 & 5 & 3 & 85.56 -- 0.04 \\
50  & 50  & 4 & 3 & 3 & 87.35 -- 0.15 \\
0   & 500 & 4 & 3 & 3 & 91.85 -- 0.05 \\
\bottomrule
\end{tabular}
\end{table}

\section{Discussion, Trade-offs, and Limitations}
\label{sec:discussion}

\paragraph{Mechanistic hypotheses.}
The clearest evidence that robustness is not explained by capacity alone comes from the $N{=}4$ family.
Across the sweep in Table~\ref{tab:hparam}, increasing oscillator dimension from $N{=}2$ to $N{=}4$ pushes clean accuracy as high as 92\% while collapsing AA-rand robustness to near zero.
This non-monotonic behavior suggests that OPL's robustness depends on a narrow dynamical regime rather than on simply increasing model flexibility.
We conjecture that the $N{=}2$ regime enforces strict degree-1 phase consensus (equivalent to a 2D rotation), creating a tightly coupled dynamical system in which individual features cannot independently align with adversarial gradient vectors. Increasing to $N{=}4$ provides enough degrees of freedom for the network to embed adversarial perturbations orthogonally to the consensus manifold, bypassing the structural defense. We predict that analogous low-dimensional couplings in other iterative architectures (such as complex-valued networks or strict 2D Lie group convolutions) should exhibit similar robustness phase transitions. Testing this hypothesis in other dynamical or iterative models is an important direction for future work.

\paragraph{Interpreting the clean--robust trade-off.}
OPL operates under a different operating regime than large adversarially trained CIFAR models~\citep{bartoldson2024adversarial,amini_meansparse_2024}.
Our best randomized checkpoint reaches 86.81\% clean accuracy and 76.63\% AA-rand robustness on CIFAR-10 without adversarial-example generation during training.
We do not interpret this as a matched-capacity or matched-compute comparison to adversarial training, since the compared methods differ in architecture, training procedure, and evaluation protocol.
Instead, the main empirical result is narrower: substantial adversarial robustness can emerge without adversarial training, but in our current setup, this robustness is associated with lower clean accuracy than the highest-capacity adversarially trained baselines.

\paragraph{Limitations.}
Our work should be considered with the following limitations in mind.
\begin{itemize}
    \item Our experiments are currently limited to CIFAR-10, CIFAR-10-C (cf. Appendix \ref{app:f_exp_res}) and CIFAR-100; and do not include higher-resolution datasets such as ImageNet. Evaluating this dataset would require implementing oscillatory neurons in model architectures that perform well in those settings, as done in the literature \cite{croce_robustbench_2021}.
    \item Although the contrast between the $N{=}2$ and $N{=}4$ regimes suggests that oscillatory dynamics matter, we do not yet provide a mechanistic theory or diagnostic that predicts in advance when a given configuration will be robust.
\end{itemize}

\section{Broader Impacts}
\label{sec:impacts}

A positive implication of this work is that robustness without adversarial training could significantly reduce the computational cost of deploying robust models, making them accessible in low-resource or on-device settings such as medical imaging or edge AI. However, the same efficiency gains could also lower the barrier for deploying robust models in surveillance, filtering, or adversarial settings where resistance to attacks may be misused. As such, improvements in robustness should be considered in the broader context of dual-use machine learning technologies.

\newpage

\bibliographystyle{plain}
\bibliography{ref}

\appendix
\newpage

\section{Mathematical Derivations: Full Kuramoto ODE, projection operators, X-PhiNet loss functions.}\label{app:Maths}

\subsection{Kuramoto layer}
The Kuramoto layers are the layers that implement the Kuramoto dynamics between oscillators and update the input image according to this dynamic. We will describe in detail here the calculations made by a Kuramoto layer in an AKOrN encoder.

A Kuramoto layer consists in a single layer of neurons that receives an initial state $X\in\mathbb{R}^{C\times H\times W}$ and an initial conditional stimuli $Y\in\mathbb{R}^{C\times H\times W}$ as input. We assume that $C = KN$, so that each spatial location $(h,w)$ of $X$ contains $K$ oscillator states $\mathbf{x}_{k,h,w}$ $\in\mathbb{R}^N$.

A single Kuramoto layer will apply $T\in\mathbb{N}^*$ discrete updates to the initial state of the oscillator system $X^{(0)}\coloneqq X$ by applying the following Kuramoto scheme. For $t=0,\dots,T-1$ :
\begin{equation}
\begin{aligned}
        &\Delta\mathbf{x}_{k,h,w}^{(t)}
=
\mathbf{\Omega}_{k,h,w}\mathbf{x}_{k,h,w}^{(t)}
+
\mathrm{Proj}_{\mathbf{x}_{k,h,w}^{(t)}}
\left(
\mathbf{y}_{k,h,w}
+
\sum_{k',\Delta h,\Delta w}
\mathbf{J}_{k,k',\Delta h,\Delta w}\,
\mathbf{x}_{k',h+\Delta h,w+\Delta w}^{(t)}
\right), \\
    &\mathbf{x}_{k,h,w}^{(t+1)} = \Pi\left(\mathbf{x}_{k,h,w}^{(t)} + \gamma \Delta\mathbf{x}_{k,h,w}^{(t)}\right)
\end{aligned}
\label{eq:akorn}
\end{equation}

where  $\mathbf{\Omega}_{k,h,w}\in\mathbb{R}^{N\times N}$ is a learned antisymmetric matrix controlling intrinsic rotation, $\Pi : \mathbf{x} \mapsto \frac{\mathbf{x}}{\lVert \mathbf{x} \rVert_2}$, $\gamma > 0$,
$\mathbf{J}_{k,k',\Delta h,\Delta w}$ are learned coupling weights,
and $\mathbf{y}_{k,h,w}$ is an input-dependent drive term. Because we use an implementation of the Kuramoto dynamical system with convolutions in our experiments, $\mathbf{J}_{k,k',\Delta h,\Delta w}$ will be the learned connectivity weights of a convolution kernel and $k, k'$ stand for output and input channels indexes.
The projection operator
\[
\mathrm{Proj}_{\mathbf{x}}(\mathbf{y}) = \mathbf{y} - \langle \mathbf{y}, \mathbf{x} \rangle \mathbf{x}
\]
ensures the velocity vector of an oscillator $\mathbf{x}$ effectively lives in the tangent plane of $\mathbf{S}^{N-1}$ at $\mathbf{x}$.

The output of this layer is the final state of the system of oscillators after the updates : $\mathbf{X}^{(T)}$.

In order to extract relevant features from this final oscillatory state, we will rely on a phase-invariant feature extractor module introduced in \cite{miyato2025artificial}.  We will motivate the use of this feature extractor below.

\subsection{Feature Extractor}\label{subsec:feature_extractor}
The feature extractor module, called the readout module $\mathcal{RM}:\mathbb{R}^{K\times N \times H\times W}\longrightarrow\mathbb{R}^{K\times N\times H\times W}$ was introduced by \cite{miyato2025artificial} and is specifically designed to leverage the symmetries and properties of the oscillators to extract meaningful features. The oscillators are constrained to live on the unit hyper-sphere of $\mathbb{R}^N$ due to the projector operator $\Pi$. As a result, all information and patterns are encoded in the \textbf{relative directions} of the oscillators rather than their absolute positions.

Consequently, the readout module $\mathcal{RM}$ is \textbf{phase-invariant}: adding a constant phase to all oscillators does not change the output of the module. Formally, for $X\in\mathbb{R}^{K\times N\times H\times W}$ and a rotation $R:\mathbb{R}^N\longrightarrow\mathbb{R}^N$, if we denote by
\[
R(X) \coloneqq \left(R(\mathbf{x}_i)\right)_{i\in[K]\times[H]\times[W]}
\]
then the readout satisfies
\[
\mathcal{RM}\big(R(\mathbf{X})\big) = \mathcal{RM}(\mathbf{X}).
\]

This formulation ensures that $\mathcal{RM}$ depends only on the \textbf{relative orientations} of the oscillators, preserving the intrinsic structure of the data encoded on the hypersphere.

In practice, we choose the simple following readout module $$\mathcal{RM}(X) = g\left(\left(\left\lVert \sum_{i}\mathbf{U}_{ji}\mathbf{x}_i\right\rVert_2\right)_{j\in[K]\times[N]\times[H]\times[W]}\right)$$ with $g:\mathbb{R}^{K\times N\times H\times W}\longrightarrow\mathbb{R}^{K\times N\times H\times W}$ a linear layer (that can be also set to identity) and $\mathbf{U}_{ji}\in\mathbb{R}^{N\times N}$ learned weight matrices.

\subsection{Oscillatory Neurons Unit}
To create a neural network from those Kuramoto layers, one needs to stack multiple Kuramoto layers and to add a readout module between two succcessive kuramoto layers that aims at extracting features from the final oscillatory states to create a new conditional stimuli $\mathbf{C}$.

\subsection{ X-PhiNet Predictive learning setup}
The X-PhiNet pre-training architecture we implemented is trained by jointly minimizing two losses : the first one, Sim-1 for the hippocampus model, and the second one, Sim-2 for the neocortex model. Those two losses are combined to update the encoder $f$ by backpropagation. \\

\paragraph{Sim-1.} This hippocampus loss is a symmetric negative cosine loss function, exactly as in the SimSiam architecture~\cite{chen_exploring_2021} :
$$
L_{\mathrm{Cos}}(\theta) = -\frac{1}{2n} \sum_{i=1}^{n}
\frac{( h^{(1)}_{i} )^\top sg({z}^{(2)}_{i})}
{\| h^{(1)}_{i} \|_{2} \, \| sg({z}^{(2)}_{i}) \|_{2}}
-\frac{1}{2n} \sum_{i=1}^{n}
\frac{sg({z}^{(1)}_{i} )^\top h^{(2)}_{i}}
{\| sg({z}^{(1)}_{i}) \|_{2} \, \| h^{(2)}_{i} \|_{2}}
$$

This loss is appropriate because it becomes minimal when the predicted representation from CA3, $h^{(1)}$, is perfectly aligned with the temporally distant signal $z^{(2)}$.\\

Next, we examine how information transfer from the hippocampus to the neocortex is captured during the model's training.

\paragraph{Sim-2.} In the neocortex, this loss is defined as an MSE loss between the CA1 output and the representation $z = f(x)$ from the entorhinal cortex (EC):

$$
L_{\mathrm{NC}}(\theta) = \frac{1}{2n} \sum_{i=1}^{n}
\left\| y^{(1)}_{i} - sg(z_{i}) \right\|_{2}^{2}
+ \frac{1}{2n} \sum_{i=1}^{n}
\left\| y^{(2)}_{i} - sg(z_{i}) \right\|_{2}^{2}
$$

The final loss for X-PhiNet is the sum of Sim-1 and Sim-2. Sim-2 is considered a slow-learning loss.

\paragraph{Stable encoder.}
To incorporate long-term memory, a different encoder $f_{long}$ has been added to X-PhiNet's third stream. This stream is designed to model the neocortex and $f_{long}$ corresponds to a stable encoder that further improve slow learning with an ability to maintain long-term signals. The weights of $f_{long}$ are updated by using an exponential moving average of the model parameters of $f$ and $f_{long}$. It is similar to techniques used by other self-supervised methods as BYOL architecture~\cite{grill_bootstrap_2020}. If we denote by $\xi$ and $\xi_{long}$ the respective parameters of $f$ and $f_{long}$ we have the following update scheme :
$$
\xi_{long} \leftarrow \beta\xi_{long} + (1-\beta)\xi
$$ with $\beta\in[0,1]$ a hyperparameter usually chosen close to 1.\\

The benefit of maintaining a separate encoder whose weights are updated via an exponential moving average (EMA) scheme is that it yields a representation network with more stable parameters during training, while simultaneously mitigating the risk of representation collapse.\\

PhiNet has demonstrated several advantages for encoder pre-training compared to other state-of-the-art architectures such as SimSiam~\cite{chen_exploring_2021} and BYOL~\cite{grill_bootstrap_2020}. Among these benefits are its \textbf{robustness to the choice of the weight decay hyperparameter} and its superior performance in online learning settings. \\

Having introduced self-supervised learning and the neuroscience-inspired models that guide our approach, we now turn to the task at hand. Our goal is to leverage the strengths of AKOrN and X-PhiNet to push beyond current state-of-the-art performance in term of robustness to adversarial attacks.

\section{Adversarial Exemples and Adversarial Attacks}\label{app:adv_attack}

While some works report only \textbf{test-set accuracies}, this evaluation alone may \textbf{not} be \textbf{sufficient} to faithfully assess the performance of a model. Indeed, it is common for a model to achieve high test accuracy while failing to learn robust and generalizable features, effectively overfitting to the data distribution. This issue becomes apparent when the model is attacked: adversarial examples---inputs that are very close to correctly classified evaluation samples---can highlight those weakness in a model, revealing its vulnerability.

\begin{figure}[h]
\centering
\includegraphics[width=0.5\linewidth]{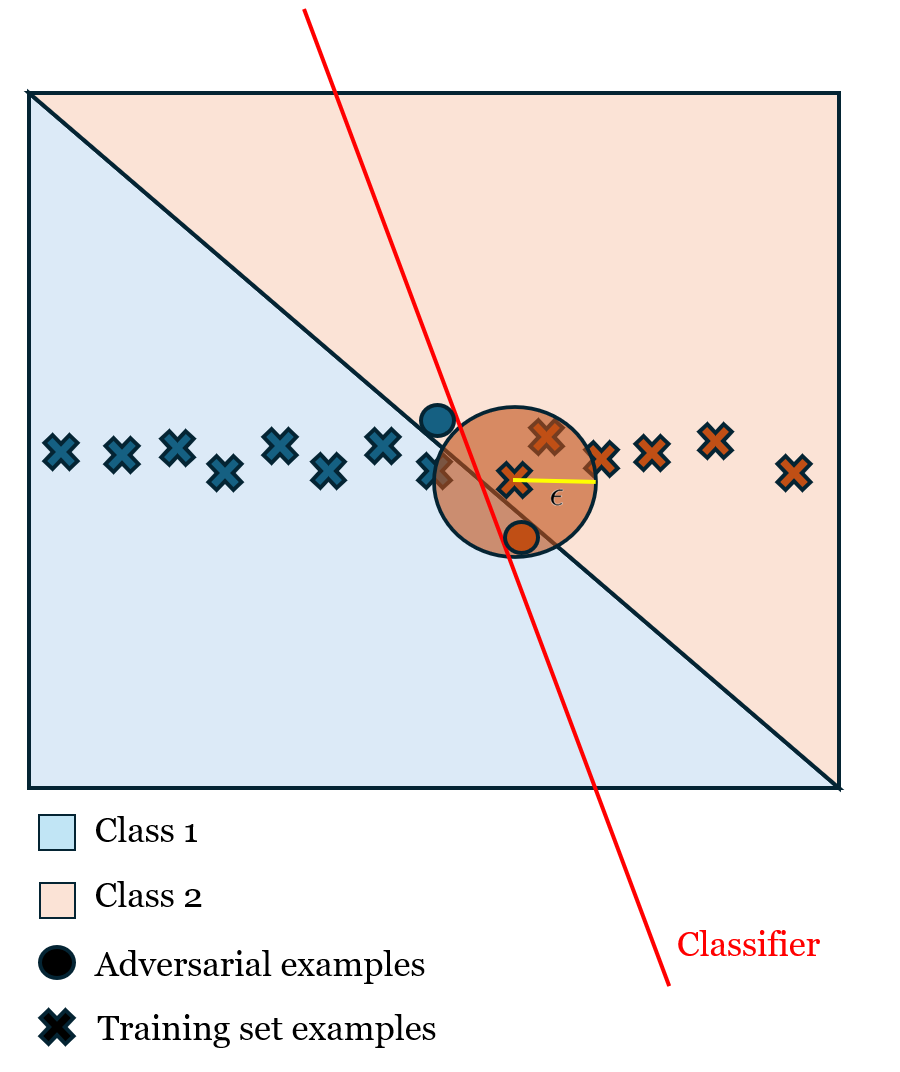}
\caption{Adversarial perturbation illustration}
\label{adversarial_perturbation_illiustration}
\end{figure}

Many different methods have been proposed to generate adversarial examples and reliably evaluate a model's robustness. Recently, one evaluation method has gained widespread popularity for its reliability. Known as \textbf{AutoAttack} \cite{croce_reliable_adv_rob_2020}, it is described by its authors as a ``parameter-free, computationally affordable, and user-independent ensemble of attacks to test adversarial robustness.'' AutoAttack has been tested on more than 50 different models and is currently the standard tool used by the reference benchmark for adversarial robustness, \textbf{RobustBench} \cite{croce_robustbench_2021}.
\\

Let's see what are the main ideas used by the AutoAttack to realize those adversarial attacks.

\subsection{Adversarial attack}

First, we formally define what constitutes an adversarial example for a $K$-class classifier
$g:\mathcal{D}\subset\mathbb{R}^d \to \mathbb{R}^K$, which assigns labels according to
$\mathrm{argmax}_k \, g_k(\cdot)$. Let $x_{\mathrm{org}} \in \mathbb{R}^d$ be a point correctly classified by $g$ as class $c$.
Given a distance metric $d(\cdot,\cdot)$ and a perturbation budget $\epsilon > 0$,
the feasible set of the attack at $x_{org}$ is defined as
\[
\mathcal{B}_\epsilon(x_{\mathrm{org}}) = \{ z \in \mathcal{D} \;|\; d(x_{\mathrm{org}}, z) < \epsilon \}.
\]

The most popular attacks rely on $l_p$ distances $d:(x,y)\in\mathbb{R}^{2d}\longmapsto\lVert x-y\rVert_p$ with $p\in\{2,\infty\}$. This choice is made for practical reasons as it will be explained below.\\
An adversarial example for $g$ at $x_{\mathrm{org}}$ is then any input such that
\[
\arg\max_{k=1,\ldots,K} g_k(z) \neq c
\quad \text{and}\quad z \in \mathcal{B}_\epsilon(x_{\mathrm{org}}).
\]
Intuitively, it is an input that is almost indistinguishable from $x_{\mathrm{org}}$
to a human observer, but is misclassified by the model.
\\
To find an adversarial example $z$, it is common to solve a constrained optimization problem
\begin{equation}\label{optim_pb_pgd}
    \max_{z \in \mathcal{D}} L(g(z), c)
\quad \text{such that} \quad
d(x_{\mathrm{orig}}, z) \leq \epsilon, \; z \in \mathcal{D}.
\end{equation}

where $L$ is a function that enforces $z$ into not being assigned to class $c$.
\newline Let's explore a popular algorithm used in adversarial attack to solve this kind of problem.

\subsubsection{Auto-PGD}
In \textbf{PGD-attack} \cite{kurakin_adv_phy_world_2016}, one of the most popular white-box attack, used by AutoAttack, the function $L$ used is the cross-entropy $-\log(q_{c}(g(z))$ where $q_{c}(g(z))$ is the predicted probability assigned to the correct class $c$. \\

To solve this optimization problem (\ref{optim_pb_pgd}), \cite{croce_reliable_adv_rob_2020} introduced \textbf{Auto-PGD}, a variant of the Projected Gradient Descet (PGD) algorithm that solves issues one may face while using PGD for solving this optimization problem. \\

Let's recall first the regular PGD algorithm. For a given $f:\mathbb{R}^d\longrightarrow\mathbb{R}, ~\mathcal{S}\subset\mathbb{R}^d$ and the problem $\max_{x\in\mathcal{S}}f(x)$, the PGD is an iterative algorithm. For $k=1,\dots,N_{iter}$,
\[
x^{(k+1)} = P_{\mathcal{S}}\left(x^{(k)}+\eta^{(k)}\nabla f(x^{(k)})\right),
\]
where $\eta^{(k)}$ is the step size at iteration $k$ and $P_{\mathcal{S}}$ is the projection onto $\mathcal{S}$ and the initial point $x^{(0)}$ is either $x_{\text{org}}$ or $x_{\text{org}}+\zeta$ with $\zeta$ a random vector such that $(x_{\text{org}}+\zeta)\in\mathcal{S}$. We can notice that Problem \ref{optim_pb_pgd} can be effectively solved for $d(\cdot,\cdot)$ being an $l_2$ or an $l_\infty$ distance.\\

In the original PGD attack, the step size is fixed, i.e. $\eta^{(k)} = \eta$ for every iteration $k$.
This choice is suboptimal and does not guarantee convergence.
In contrast, Auto-PGD divides the total number of iterations $N_{\text{iter}}$ into an \textbf{exploration phase} and an \textbf{exploitation phase},
during which the step size $\eta$ is adaptively updated according to different criteria.
 \\

 Let's present the ideas of Auto-PGD in four steps.

\paragraph{Gradient step.}
Auto-PGD updates use a step size $\eta^{(k)}$ at iteration $k$ and incorporates a momentum term, which is absent in the original PGD algorithm. The initial steps of APGD are typically large, so the momentum helps to regularize the updates by incorporating information from previous iterations, smoothing the optimization trajectory. The update step corresponds to the following :

\begin{align*}
z^{(k+1)} &= P_{S}\left(x^{(k)} + \eta^{(k)}\nabla f(x^{(k)})\right) \\
x^{(k+1)} &= P_{S}\left(x^{(k)} + \alpha \cdot (z^{(k+1)} - x^{(k)}) + (1 - \alpha) \cdot (x^{(k)} - x^{(k-1)})\right)
\end{align*}
where $\alpha\in[0,1]$ regulates the influence of the momentum term. $\alpha = 0.75$ is used in practice.

\paragraph{Step size selection.}
The initial step size at iteration 0 is chosen such that $\eta^{(0)}=2\epsilon$. Then the algorithm will decide weather or not the step size will be halved at fixed chosen checkpoints $w_0=0<w_1,\dots<w_n<N_{\text{iter}}$. To update the step size, one of the two following conditions has to be true:
\begin{align*}
 1. & \quad \sum_{i=w_{j-1}}^{w_j-1} \mathbf{1}_{f(x^{(i+1)}) > f(x^{(i)})} < \rho \cdot (w_j - w_{j-1}), \\
 2. & \quad \eta^{(w_{j-1})} \equiv \eta^{(w_j)} \quad \text{and} \quad f_{\max}^{(w_{j-1})} \equiv f_{\max}^{(w_j)},
\end{align*}

\begin{itemize}
    \item \textit{1. Insufficient Improvement:} The step size is halved if the proportion of iterations that increased the objective function $f$ since the prior checkpoint, $w_{j-1}$, is below a threshold $\rho$ (typically $\rho = 0.75$).

    \item \textit{2. Stagnation and Cycling Prevention:} The step size is also halved if it remained unchanged at the previous checkpoint ($w_{j-1}$) \textbf{and} the maximum value of the objective function $f$ has not improved since that checkpoint. This rule helps the algorithm escape cycles.
\end{itemize}

\paragraph{Checkpoints restart.}
At each checkpoint $w_j$ where the step size has been halved, the algorithm restarts from the best point found so far. It means in that case that we fix $x^{w_{j}+1}\coloneqq x_{\text{max}}$ with $x_{\text{max}}\coloneqq \textrm{argmax}_{k\leq w_j}f(x^{(k)})$. The idea is to refine the research of an adversarial example around the neighborhood of the current best candidate solution.

\paragraph{Exploration vs Exploitation.}
The choice of the checkpoints $w_j$ at which the step size can be reduced is important to define a transition between an initial exploration phase exploring the whole feasible set $\mathcal{S}$ and an exploitation phase where the step size is reduced more often leading to more frequent improvements of the objective but with smaller magnitude. \\

The checkpoints are defined as follow:

$$
\begin{aligned}
    &w_j \coloneqq \lceil p_j N_{\text{iter}}\rceil \le N_{\text{iter}}
\end{aligned}
$$
with $p_j\in[0,1]$ defined as
$$\left\{
\begin{aligned}
    &p_0=0 \\
    &p_1=0.22 \\
    &p_{j+1} = p_j + \max\{p_j-p_{j-1}-0.03,0.06\}
\end{aligned}\right.
$$
The only remaining free parameter in the implementation of AutoAttack is the budget $N_{\text{iter}}$. By construction, the resulting checkpoints are increasingly dense as the iterations progress.\\

In conclusion, the ensemble AutoAttack consists in combination of two parameter-free versions of PGD, APGD with cross-entropy loss, APGD with Difference of Logits Ratio Loss \cite{croce_reliable_adv_rob_2020}, and two existing complementary attacks, FAB \cite{croce2020minimally_fab} and Square Attack \cite{andriushchenko2020square}.\\

The \textbf{adversarial accuracy} of a model is obtained by evaluating it under an adversarial attack. It corresponds to the number of correctly classified samples that cannot be successfully perturbed into adversarial examples by the attacker (here, AutoAttack) divided by the number of samples in total. It is this adversarial accuracy that serves as the principal metric highlighted in robustness benchmarks for comparing model robustness.\\

In the next part, we will focus on the current state of the research in adversarial accuracy. We will rely on RobustBench \cite{croce_robustbench_2021}, a standardized online benchmark for adversarial robustness.

\section{Experimental Settings and Reproducibility Details}\label{app:exp_sett}

All experiments follow a two-stage training pipeline consisting of self-supervised pretraining followed by supervised fine-tuning. Unless otherwise stated, the results reported for the OPL CIFAR-10 setting use the configuration described below.

\paragraph{Training pipeline.}
We first pretrain the model using a self-supervised objective and then fine-tune the pretrained backbone for downstream image classification. For the experiments in the main text, the self-supervised method is \textbf{PhiNet} and the backbone is \textbf{AKOrN}. Fine-tuning is initialized from the pretrained checkpoint.

\paragraph{Dataset and data loading.}
We use \textbf{CIFAR-10}, which contains 10 classes. The default data root is \texttt{data/}. During training, dataloaders use shuffling for the training split and no shuffling for the test split. All dataloaders use \texttt{num\_workers=1} and \texttt{pin\_memory=True}.

For pretraining, augmentations are determined by the SSL method, and in the X-PhiNet setting we use the \textbf{PhiNet augmentation pipeline}. For fine-tuning, we use the \texttt{finetuning} augmentation strategy implemented by \texttt{augmentation\_strong(...)}. For evaluation, we use a minimal test-time transform consisting of \texttt{ToTensor()} only.

\paragraph{AutoAttack.}\label{autoattack_full_explaination}
We use AutoAttack to compare with the state-of-the-art adversarial purification methods. To make a fair comparison, we uses their codebase: \href{https://github.com/RobustBench/}{https://github.com/RobustBench/} with default hyperparameters for evaluation. Similarly, we set $\epsilon = 8/255$ for AutoAttack $l_{\infty}$, on CIFAR-10 and CIFAR-100.

There are two versions of AutoAttack: (i) the Standard version, which contains four attacks: APGD-CE, APGD-T, FAB-T and Square, and is mainly used for evaluating deterministic defense methods, and (ii) the Rand version, which contains two attacks: APGD-CE and APGD-DLR, and is used for evaluating stochastic defense methods. Because our method is stochastic, we use the Rand version and set EoT=20 by default.

\paragraph{Optimization.}
Both pretraining and fine-tuning use the \textbf{Adam} optimizer. Pretraining is run for \textbf{700 epochs} with batch size \textbf{512}, learning rate \textbf{$1\times 10^{-4}$}, and weight decay \textbf{$1\times 10^{-5}$}. Fine-tuning is run for \textbf{400 epochs} with batch size \textbf{512}, learning rate \textbf{$1\times 10^{-4}$}, and weight decay \textbf{0.0}. For single-run ablations, we use seed 0. For the main CIFAR-10 result, we report the mean and standard deviation over five independently trained runs. Training is run on \textbf{CUDA if available}, otherwise CPU.

\paragraph{PhiNet configuration.}
For X-PhiNet, we use a projection output dimension of \textbf{2048}. The X-PhiNet loss uses an \textbf{MSE loss ratio of 0.5}, an \textbf{orientation loss ratio of 0.1}, and an \textbf{EMA coefficient $\beta = 0.99$} for the slow encoder update.

\paragraph{AKOrN backbone configuration.}
For the AKOrN backbone, we use \textbf{128 channels}, oscillator dimensionality \textbf{$n=2$}, \textbf{$T=3$} recurrent steps, and \textbf{$L=3$} layers. The interaction operator is convolutional (\texttt{J="conv"}), with kernel sizes \textbf{[9, 7, 5]}. We use reorientation kernel size \textbf{3} and reorientation order \textbf{2}. Normalization settings are \textbf{batch normalization} (\texttt{norm="bn"}) and \textbf{group normalization} (\texttt{c\_norm="gn"}). Additional AKOrN settings are \texttt{gamma=1.0}, \texttt{use\_omega=True}, \texttt{init\_omg=1.0}, \texttt{global\_omg=True}, \texttt{learn\_omg=True}, and \texttt{ensemble=1}. Randomness in the AKOrN forward pass is enabled.

\paragraph{Checkpointing and logging.}
During pretraining, if the SSL model exposes a slow encoder, we save the slow encoder backbone; otherwise, we save the backbone weights directly. During fine-tuning, we save both the final checkpoint.

\paragraph{Exact configuration used.}
Table~\ref{tab:repro_settings} summarizes the exact hyperparameter configuration corresponding to the training script used in our experiments.

\begin{table}[h]
\centering
\small
\caption{Experimental settings for reproducibility for the OPL CIFAR-10 experiments.}
\begin{tabularx}{\textwidth}{@{}llX@{}}
\toprule
Category & Setting & Value \\
\midrule
Training pipeline & Phases & Self-supervised pretraining + supervised fine-tuning \\
SSL method & Method & X-PhiNet \\
Backbone & Architecture & AKOrN \\
Dataset & Dataset & CIFAR-10 \\
Dataset & Number of classes & 10 \\
Random seed & Seed & 0 \\
Device & Hardware selection & CUDA if available, else CPU \\
Pretraining & Epochs & 400 \\
Fine-tuning & Epochs & 400 \\
Pretraining & Batch size & 512 \\
Fine-tuning & Batch size & 512 \\
Pretraining & Optimizer & Adam \\
Fine-tuning & Optimizer & Adam \\
Pretraining & Learning rate & $1\times 10^{-4}$ \\
Fine-tuning & Learning rate & $1\times 10^{-4}$ \\
Pretraining & Weight decay & $1\times 10^{-5}$ \\
Fine-tuning & Weight decay & 0.0 \\
SSL head & Projection dimension & 2048 \\
AKOrN & Channels ($ch$) & 128 \\
AKOrN & Oscillator dimensions ($n$) & 2 \\
AKOrN & Time steps ($T$) & 3 \\
AKOrN & Number of layers ($L$) & 3 \\
AKOrN & Interaction operator ($J$) & conv \\
AKOrN & Kernel sizes & [9, 7, 5] \\
AKOrN & Reorientation kernel size & 3 \\
AKOrN & Reorientation order ($ro_N$) & 2 \\
AKOrN & Normalization & bn \\
AKOrN & Channel normalization & gn \\
AKOrN & Gamma & 1.0 \\
AKOrN & Use omega & True \\
AKOrN & Initial omega & 1.0 \\
AKOrN & Global omega & True \\
AKOrN & Learn omega & True \\
AKOrN & Ensemble size & 1 \\
AKOrN & Randomness in forward pass & True \\
PhiNet & MSE loss ratio & 0.5 \\
PhiNet & Orientation loss ratio & 0.1 \\
PhiNet & EMA beta & 0.99 \\
Pretraining augmentation & Strategy & X-PhiNet \\
Fine-tuning augmentation & Strategy & finetuning \\
Evaluation augmentation & Test transform & \texttt{ToTensor()} \\
Data loading & Train shuffle & True \\
Data loading & Test shuffle & False \\
Data loading & Num. workers & 1 \\
Data loading & Pin memory & True \\
Logging & W\&B & Enabled \\
Logging & TensorBoard & Enabled for fine-tuning \\
Checkpointing & Pretraining save target & Slow encoder if available; otherwise backbone \\
Checkpointing & Fine-tuning save target & Last and best checkpoints \\
\bottomrule
\end{tabularx}

\label{tab:repro_settings}
\end{table}

\paragraph{Command used.}
For completeness, the following command was used for the reported configuration:

\begin{verbatim}
python3 train.py \
  --dataset cifar10 \
  --run_pretraining \
  --run_finetuning \
  --load_from_pretrain \
  --ssl_method X-PhiNet \
  --backbone akorn \
  --pretrain_epochs 400 \
  --finetune_epochs 400 \
  --pretrain_bs 512 \
  --finetune_bs 512 \
  --pretrain_lr 1e-4 \
  --finetune_lr 1e-4 \
  --out_dim 2048 \
  --ch 128 \
  --randomness True \
  --n 2 \
  --T 3 \
  --L 3 \
  --use_wandb \
  --mse_loss_ratio 0.5 \
  --ori_loss_ratio 0.1 \
  --pretrain_weight_decay 1e-5
\end{verbatim}
\section{Further Experimental Results}\label{app:f_exp_res}

\paragraph{OPL with different predictive learning Frameworks }
Table \ref{tab:ssl_comparison} displays initial experiments conducted with different predictive learning frameworks before we selected X-PhiNet for later experiments and work.
\begin{table}[h]
\centering
\caption{Adversarial attack robustness on CIFAR-10 for different predictive learning frameworks.}
\begin{tabular}{lcc|cc}
\toprule
\multirow{2}{*}{Method} & \multicolumn{2}{c|}{Pretraining: 200 epochs} & \multicolumn{2}{c}{Pretraining: 400 epochs} \\
\cmidrule(r){2-3} \cmidrule(l){4-5}
 & Clean Acc & Robust Acc & Clean Acc & Robust Acc \\
\midrule
BYOL    & 82.71 & 58.57 & 76.05 & 46.15 \\
SimSiam & 84.22 & \textbf{68.89} & \textbf{83.50} & 74.80 \\
Phinet  & \textbf{84.36} & 68.79 & 84.35 & \textbf{76.56} \\
\bottomrule
\end{tabular}
\label{tab:ssl_comparison}
\end{table}
\paragraph{Training compute.}
Table~\ref{tab:compute_comparison_scalinglaw} reports absolute training FLOPs for OPL and representative adversarially trained reference models. FLOPs for OPL and AKOrN are estimated using a PyTorch FLOP counter.
Because OPL does not generate adversarial examples during training, \textbf{its overall training cost is 358 times lower than the current SOTA models}. Results are not directly comparable as the adversarially trained models are evaluated under standard AutoAttack.

\begin{table}[h]
\centering
\small
\caption{
\textbf{Robustness to adversarial examples by AutoAttack and absolute training-cost.}
Models marked with $^\ast$ use stochastic forward passes and are evaluated with AutoAttack-rand (EoT). The top two methods are selected from the highestranked methods on \href{https://robustbench.github.io/}{https://robustbench.github.io/}
}
\label{tab:compute_comparison_scalinglaw}
\begin{tabular}{lccc}
\toprule
\textbf{Model} & \textbf{Reported robust acc.} & \textbf{Protocol} & \textbf{Training FLOPs} \\
\midrule
Bartoldson et al. (2024) \cite{bartoldson2024adversarial} & 73.71 & standard AA & $1.43 \times 10^{21}$ \\
Amini et al. (2024) \cite{amini_meansparse_2024} & 75.28 & standard AA & $> 10^{21}$ \\
AKOrN$^\ast$ \cite{miyato2025artificial}& 64.98 & AA-rand, EoT $K=20$ & $1.19 \times 10^{18}$  \\
\midrule
\textbf{OPL (ours)$^\ast$} & \textbf{76.63 $\pm$ 0.76} & AA-rand, EoT $K=20$ & $\mathbf{3.99 \times 10^{18}}$ \\
\bottomrule
\end{tabular}

\end{table}

\begin{table}
\centering
\caption{
CIFAR-10-C robustness. Accuracy is averaged over the five severity levels
for each corruption type. OPL improves over AKOrN for every corruption.
}
\label{tab:cifar10c_corruptions}
\begin{tabular}{lccc}
\toprule
Corruption & AKOrN & OPL & Gain \\
\midrule
Brightness & 83.81 & \textbf{86.20} & +2.39 \\
Contrast & 78.36 & \textbf{80.80} & +2.44 \\
Elastic Transform & 79.94 & \textbf{82.02} & +2.08 \\
Pixelate & 83.59 & \textbf{85.90} & +2.31 \\
JPEG Compression & 83.12 & \textbf{84.74} & +1.62 \\
Gaussian Noise & 82.12 & \textbf{82.86} & +0.74 \\
Shot Noise & 82.94 & \textbf{83.88} & +0.94 \\
Impulse Noise & 78.16 & \textbf{78.59} & +0.43 \\
Defocus Blur & 82.93 & \textbf{84.79} & +1.86 \\
Glass Blur & 77.47 & \textbf{79.30} & +1.83 \\
Motion Blur & 79.69 & \textbf{81.37} & +1.68 \\
Zoom Blur & 83.59 & \textbf{86.06} & +2.47 \\
Snow & 78.38 & \textbf{80.83} & +2.45 \\
Frost & 80.71 & \textbf{84.19} & +3.48 \\
Fog & 75.60 & \textbf{77.59} & +1.99 \\
\midrule
Mean & 80.69 & \textbf{82.61} & +1.92 \\
\bottomrule
\end{tabular}
\end{table}
\paragraph{CIFAR-10-C robustness.}
We further evaluate robustness to natural distribution shifts using
CIFAR-10-C~\cite{daniel_hendrycks_2019_2535967}, which contains
15 corruption types at five severity levels. Table~\ref{tab:cifar10c_corruptions}
reports corruption-wise accuracy averaged over severities. OPL improves the
mean corruption accuracy from 80.69\% to 82.61\%, yielding a +1.92 percentage
point gain over AKOrN. Equivalently, OPL reduces the mean corruption error
from 19.31\% to 17.39\%, corresponding to a 9.9\% relative error reduction.
Notably, the improvement is consistent across all 15 corruption types,
including noise, blur, weather, and digital artifacts. These results indicate
that the robustness benefits of OPL are not limited to adversarial
perturbations, but also transfer to common natural corruptions.

\paragraph{Extended hyperparameter sweeps.}
To better understand the sensitivity of OPL to optimization and architectural choices, we performed additional hyperparameter sweeps over the use of AKOrN forward-pass randomness, the oscillator dimensionality $N$, the pretraining duration, the X-PhiNet loss weights, the EMA coefficient $\beta$, and pretraining weight decay. Unless otherwise stated, all runs use the same base setting as the main experiments: CIFAR-10, AKOrN backbone, $ch=128$, $T=3$, $L=3$, 400 fine-tuning epochs, and the same evaluation protocol as in the main text.

Table~\ref{tab:extra_sweeps} summarizes these exploratory runs. In the table, \textbf{C-R} reports the pair \emph{clean accuracy / robust accuracy}. Here, robust accuracy denotes the adversarial accuracy under the robustness evaluation protocol used in the main paper. When the final column contains entries such as \texttt{0.99 / 1e-5}, the first value denotes the EMA coefficient $\beta$ and the second denotes the \emph{pretraining weight decay}. Thus, \texttt{1e-5} refers to pretraining weight decay, not to $\beta$.

\paragraph{Effect of randomness and oscillator dimensionality.}
We first studied the effect of enabling randomness in the AKOrN forward pass and varying the oscillator dimensionality $N$. Increasing $N$ from 2 to 4 while keeping randomness enabled led to a severe collapse in robust accuracy despite strong clean accuracy, indicating that this setting is unstable from a robustness perspective. In contrast, disabling randomness with $N=2$ yielded a much stronger clean/robust tradeoff than the corresponding randomized setting with orientation loss disabled.

\paragraph{Effect of X-PhiNet loss weights.}
We also varied the MSE and orientation loss ratios used in X-PhiNet pretraining. Across these runs, moderate MSE weighting produced the strongest overall tradeoff, while adding a nonzero orientation term generally improved robustness relative to the zero-orientation baseline with randomness enabled. However, the results also indicate some sensitivity to the precise weighting, suggesting that the method benefits from careful balancing of representation alignment and orientation regularization.

\paragraph{Effect of pretraining duration.}
Increasing pretraining from 400 to 700 epochs improved robust accuracy in our sweep while preserving strong clean performance. This suggests that longer self-supervised training can continue to improve the learned representation for downstream robustness, although the gains are not strictly monotonic across all settings.

\paragraph{Effect of EMA coefficient and pretraining weight decay.}
We additionally explored the role of the EMA coefficient $\beta$ and pretraining weight decay. In these experiments, the notation \texttt{$\beta$ / wd} is used, where \texttt{wd} denotes the pretraining weight decay. We found that introducing a small amount of pretraining weight decay (e.g., $10^{-5}$) was often beneficial for stabilizing training, while the best-performing runs continued to use $\beta$ values close to 1.0. These observations motivated the default setting used in our main experiments.

\begin{table}[h]
\centering
\small
\caption{Extended hyperparameter sweeps for X-PhiNet+AKOrN on CIFAR-10. C-R denotes clean accuracy / robust accuracy. Pret. wd denotes pretraining weight decay.}
\label{tab:extra_sweeps}
\begin{tabular}{cccccccccccc}
\toprule
Rand & Pret. & Fint. & $ch$ & $N$ & $T$ & $L$ & C-R (\%) & MSE & Ori & $\beta$ & Pret. wd \\
\midrule
yes  & 400 & 400 & 128 & 4 & 3 & 3 & 88.65 / 0.62  & 0.5  & 0.0 & 0.995 & 0 \\
no   & 400 & 400 & 128 & 2 & 3 & 3 & 85.25 / 76.81 & 0.5  & 0.0 & 0.995 & 0 \\
yes  & 400 & 400 & 128 & 2 & 3 & 3 & 84.60 / 76.08 & 0.25 & 0.1 & 0.99  & 0 \\
yes  & 400 & 400 & 128 & 2 & 3 & 3 & 87.20 / 74.20 & 0.5  & 0.1 & 0.99  & 1e-5 \\
yes  & 700 & 400 & 128 & 2 & 3 & 3 & 86.81 / 77.66 & 0.5  & 0.1 & 0.99  & 1e-5 \\
yes  & 400 & 400 & 128 & 2 & 3 & 3 & 87.91 / 75.85 & 0.25 & 0.1 & 0.99  & 1e-5 \\
yes  & 400 & 400 & 128 & 2 & 3 & 3 & 87.79 / 75.72 & 0.1  & 0.1 & 0.99  & 1e-5 \\
yes  & 400 & 400 & 128 & 2 & 3 & 3 & 89.61 / 30.13 & 0.25 & 0.1 & 0.99  & 1e-5 \\
\bottomrule
\end{tabular}

\end{table}
\subsection{Compute resources}
\label{app:compute_resources}

All experiments reported in this paper were conducted on a single NVIDIA RTX A6000 GPU with 48GB of GPU memory. We did not use multi-GPU training, TPUs, or distributed execution. CPU resources were used only for standard data loading, preprocessing, checkpointing, and logging. All reported experiments use CIFAR-scale datasets, so storage requirements are modest and dominated by saved checkpoints and logging files rather than by dataset size. Robustness evaluations were run on the same single-GPU setup. EoT-based attacks are substantially more expensive than clean evaluation because they require multiple stochastic forward/backward passes per attack step.

Table~\ref{tab:compute_resources} summarizes the compute resources used for the main experiment classes. Training FLOPs are estimated with a PyTorch FLOP counter and include the reported training stages, but exclude logging overhead, checkpoint I/O, preliminary failed runs, and adversarial evaluation cost. The full research project required additional compute for exploratory sweeps over oscillator dimension, randomness, SSL objective, pretraining length, loss weights, EMA coefficient, and weight decay; these exploratory runs are reported separately in Appendix~\ref{app:f_exp_res}.

\begin{table}[h]
\centering
\small
\caption{Compute resources used for the main experiment classes. All experiments were run on a single NVIDIA RTX A6000 GPU with 48GB of memory. FLOP estimates exclude logging, checkpoint I/O, preliminary failed runs, and adversarial evaluation cost.}
\label{tab:compute_resources}
\begin{tabularx}{\textwidth}{@{}l c X X@{}}
\toprule
Experiment class & GPUs & Training schedule & Estimated training FLOPs \\
\midrule
AKOrN baseline
& 1
& Supervised training on CIFAR-10.
& \(1.19 \times 10^{18}\) \\

OPL main run
& 1
& X-PhiNet self-supervised pretraining followed by supervised fine-tuning.
& \(3.99 \times 10^{18}\) \\

SSL objective ablations
& 1
& Same AKOrN backbone with alternative predictive SSL objectives.
& Not separately recorded \\

Hyperparameter sweeps
& 1
& Sweeps over oscillator dimension, randomness, pretraining length, loss weights, EMA coefficient, and weight decay.
& Not separately recorded \\

CIFAR-10-C evaluation
& 1
& Evaluation only; no additional training.
& Not applicable \\

Adversarial evaluation
& 1
& AutoAttack-rand with EoT, PGD with EoT, Square Attack, and transfer attacks.
& Not included in training FLOPs \\
\bottomrule
\end{tabularx}
\end{table}

We did not systematically record wall-clock time for every exploratory run. We therefore report hardware type, GPU memory, number of GPUs, training schedules, and estimated FLOPs, but not a complete wall-clock accounting for every experiment.

\subsection{Code Availability}
\label{app:code}

For reproducibility, we provide an anonymous repository containing the OPL pipeline and the training and evaluation scripts used in this work:
\url{https://anonymous.4open.science/r/OPL-72D2}

\end{document}